\documentclass[runningheads]{llncs}
\usepackage[T1]{fontenc}
\usepackage{graphicx}
\usepackage{amssymb}
\usepackage{amsmath}
\usepackage{multirow}
\usepackage{booktabs}
\usepackage{bm}
\usepackage{cite}
\usepackage[colorlinks=true,linkcolor=blue,citecolor=blue,urlcolor=blue,]{hyperref}
\usepackage{marvosym}

\begin{document}
\title{UniCrossAdapter: Multimodal Adaptation of CLIP for Radiology Report Generation}
\titlerunning{UniCrossAdapter}
%
\author{Yaxiong Chen\inst{1,2} \and 
Chuang Du\inst{1}\thanks{Work done during an internship at MedAI Technology (Wuxi) Co. Ltd.} \and
Chunlei Li\inst{3} \and
Jingliang Hu\inst{3} \and
Yilei Shi\inst{3} \and
Shengwu Xiong\inst{1,2} \and
Xiao Xiang Zhu\inst{4} \and
Lichao Mou\inst{3}\textsuperscript{(\Letter)}}


%
\authorrunning{Y. Chen et al.}
%
\institute{Wuhan University of Technology, Wuhan, China \and Shanghai Artificial Intelligence Laboratory, Shanghai, China \and MedAI Technology (Wuxi) Co. Ltd., Wuxi, China\\\email{lichao.mou@medimagingai.com} \and Technical University of Munich, Munich, Germany}
\maketitle              
\begin{abstract}
Automated radiology report generation aims to expedite the tedious and error-prone reporting process for radiologists. While recent works have made progress, learning to align medical images and textual findings remains challenging due to the relative scarcity of labeled medical data. For example, datasets for this task are much smaller than those used for image captioning in computer vision. In this work, we propose to transfer representations from CLIP, a large-scale pre-trained vision-language model, to better capture cross-modal semantics between images and texts. However, directly applying CLIP is suboptimal due to the domain gap between natural images and radiology. To enable efficient adaptation, we introduce UniCrossAdapter, lightweight adapter modules that are incorporated into CLIP and fine-tuned on the target task while keeping base parameters fixed. The adapters are distributed across modalities and their interaction to enhance vision-language alignment. Experiments on two public datasets demonstrate the effectiveness of our approach, advancing state-of-the-art in radiology report generation. The proposed transfer learning framework provides a means of harnessing semantic knowledge from large-scale pre-trained models to tackle data-scarce medical vision-language tasks. Code is available at \url{https://github.com/chauncey-tow/MRG-CLIP}.

\end{abstract}
\section{Introduction}
Radiology report writing is a tedious and error-prone task for radiologists due to the large volume of images needing interpretation. Automated report generation has recently emerged as a promising solution to expedite this process and alleviate the workload for radiologists. This task bears similarity to image captioning in computer vision, whereby textual descriptions must be produced to characterize visual inputs.
\par
There has been growing interest in this direction. The authors of~\cite{ref_1} propose to generate radiology reports with a memory-driven Transformer and firstly conduct studies on MIMIC-CXR dataset~\cite{ref_2}. They later augment their model with a cross-modal memory module~\cite{ref_3}. \cite{ref_4} puts forth an approach to distill both posterior and prior knowledge to further boost performance. In order to better align visual and textual features, \cite{ref_5} employs reinforcement learning over the cross-modal memory network~\cite{ref_3}. In~\cite{ref_6}, the authors design a cross-modal prototype network to facilitate interactions across modalities. Aiming to promote semantic alignment, \cite{ref_7} explicitly leverage text embeddings to guide visual feature learning. Recently, \cite{ref_8} introduces a framework that makes use of a dynamic graph to enhance visual representations in a contrastive learning paradigm for radiology report generation tasks.
\par
Due to medical privacy concerns, the difficulty of gathering medical data, and the labor-intensive nature of annotation, the amount of data available for radiology report generation is relatively small compared to that used for image captioning in computer vision. For example, IU-Xray (4K images)~\cite{ref_9} and MIMIC-CXR (220K images)~\cite{ref_2} are much smaller than image captioning datasets Conceptual Captions (3.3M images)~\cite{ref_10} and Conceptual 12M (12M images)~\cite{ref_11}. Learning comprehensively from such limited data makes it challenging for current methods to fully understand cross-modal semantics between radiological images and reports~\cite{ref_1,ref_3,ref_4,ref_5,ref_6,ref_7,ref_8}. Overcoming this paucity of labeled data to better learn these semantics is crucial for advancing radiology report generation.
\par
Recently, leveraging large-scale pre-trained vision-language models, such as CLIP~\cite{ref_12}, which is trained on 400 million image-text pairs collected from the internet to match images with their corresponding textual descriptions, has become a promising approach for tackling downstream tasks in computer vision. However, the application of such models on radiology report generation still remains unexplored. In this work, we propose transferring the knowledge encapsulated in CLIP to the task of automatic report generation to better model the semantic relationship between medical images and their associated radiological findings.

Despite its strong performance, directly applying CLIP to radiology report generation tasks poses certain challenges. CLIP has been pre-trained on large-scale natural image-text datasets, exhibiting a substantial domain divergence from medical images. Therefore, while the model encapsulates rich semantic knowledge about everyday scenes, fine-tuning is imperative to adapt CLIP to radiology. However, conducting a full fine-tuning of a model as massive as CLIP is highly impractical given immense computational demands. To enable efficient adaptation, we propose uni- and cross-modal adapter (UniCrossAdapter), a parameter-efficient fine-tuning approach to adapt CLIP for the task of radiology report generation. The key idea is to integrate lightweight adapter modules into CLIP that can be fine-tuned on the target task while keeping the pre-trained backbone parameters frozen. The modules are distributed to both visual and textual modalities and their interactions for better aligning medical images and texts. Our contributions are three-fold.
\begin{itemize}
    \item We investigate the transfer of representations learned by CLIP to describe medical image findings.
    \item We introduce a novel adapter architecture that improves vision-language alignment on radiology images and reports by coupling image and text adapter modules through a cross-attention mechanism.
    \item Our approach achieves state-of-the-art performance on IU-Xray and MIMIC-CXR, the two most used benchmark datasets.
\end{itemize}

\section{Method}
\begin{figure}[t]
\begin{center}
\includegraphics[width=\linewidth]{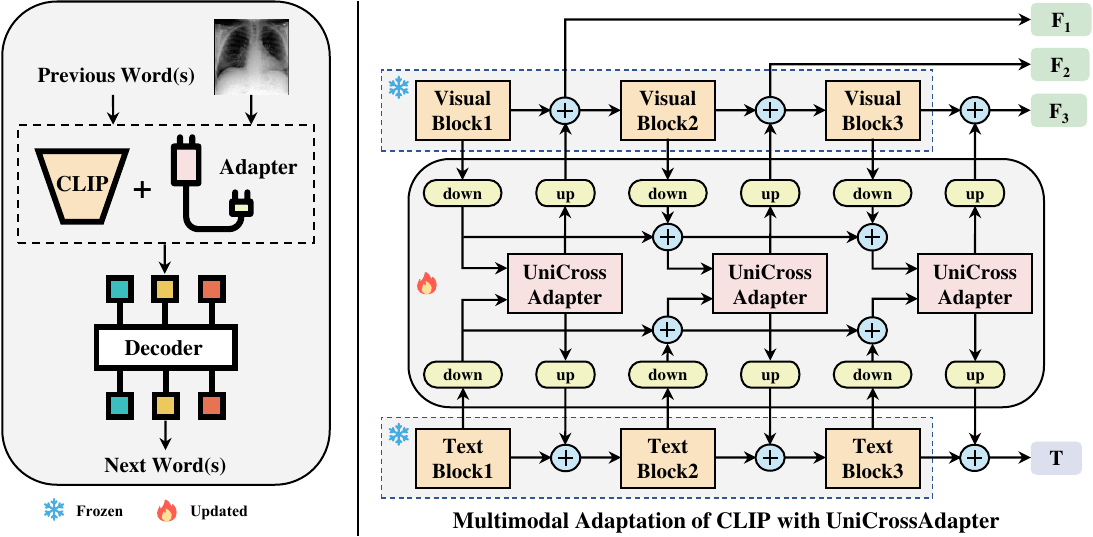}
\renewcommand{\figurename}{Fig.}
\end{center}
    \caption{(Left) Overall architecture of our method for radiology report generation, leveraging CLIP and the proposed UniCrossAdapter. (Right) Illustration of the interaction between the UniCrossAdapter and CLIP's text and image encoders.}
\label{fig:main_fig}
\end{figure} 

We propose an end-to-end framework for automatic radiology report generation, as illustrated in Fig.~\ref{fig:main_fig}. The model comprises two key components: (i) the adaptation of CLIP with UniCrossAdapter to learn visual and textual representations for radiology data, and (ii) a decoder that generates reports. In what follows, we first detail each of them. Then, we describe the training and inference procedures.
\par

\subsection{Multimodal Adaptation of CLIP with UniCrossAdapter}
Recent work has explored parameter-efficient fine-tuning methods~\cite{ref_13,ref_14,ref_15,ref_17,ref_18,ref_19,ref_20} for adapting large pre-trained models to downstream tasks. However, architectures used in prior efficient tuning techniques, e.g., down-up feedforward layers~\cite{ref_19,ref_20} and LoRA~\cite{ref_18}, may be too simple to effectively adapt complex multimodal models. Moreover, most existing approaches have focused largely on unimodal or basic classification tasks, with little exploration on more challenging multimodal setups requiring inter-modality interaction modeling. Our proposed UniCrossAdapter is dedicated to the multimodal adaptation of CLIP.

\subsubsection{CLIP's Text and Image Encoders}
We utilize the pre-trained CLIP text Transformer to extract text features. Due to its large parameter size, the text Transformer remains frozen during fine-tuning. We then evenly split it into three sequential blocks and denote the text feature map from each block as $\bm{T}_i\in\mathbb{R}^{N\times D}$, where $i\in\{1,2,3\}$, $N$ is the number of tokens, and $D$ is the feature dimension.
\par
For the visual branch, we use CLIP's image encoder, specifically ResNet-101, to extract multi-scale visual features $\bm{F}_{i}$ from the last three stages. Similar to the text encoder, we freeze the image encoder during fine-tuning to leverage rich semantics learned from pre-training.

\subsubsection{Unimodal and Cross-Modal Adaptation}
The visual and linguistic features are first projected to a lower-dimensional space. Residual connections are further formed  between consecutive adapter layers to enrich unimodal representations. This process can be formulated as
\begin{equation}
\begin{split}
\hat{\bm{F}}_i&=\mathrm{down}(\bm{F}_i)+\hat{\bm{F}}_{i-1} \,, \\
\hat{\bm{T}}_i&=\mathrm{down}(\bm{T}_i)+\hat{\bm{T}}_{i-1} \,,
\end{split}
\end{equation}
where $\mathrm{down}(\cdot)$ indicates dimension reduction layers implemented by convolutional and linear layers for visual and textual features, respectively. To encourage interactions within each modality, we apply multi-head self-attention (MHSA) on both modalities:
\begin{equation}
\begin{split}
\bm{F}_i^{sa}&=\mathrm{MHSA}(\hat{\bm{F}}_i) \,, \\
\bm{T}_i^{sa}&=\mathrm{MHSA}(\hat{\bm{T}}_i) \,.
\end{split}
\end{equation}
For coupling the visual and linguistic adapter modules, we perform multi-head cross-attention (MHCA) across the adapted unimodal representations for establishing cross-modal interactions:
\begin{equation}
\begin{split}
\bm{F}_i^{ca}&=\mathrm{FFN}(\mathrm{MHCA}(Q=\bm{F}_i^{sa},K=\hat{\bm{T}}_i,V=\hat{\bm{T}}_i)) \,, \\
\bm{T}_i^{ca}&=\mathrm{FFN}(\mathrm{MHCA}(Q=\bm{T}_i^{sa},K=\hat{\bm{F}}_i,V=\hat{\bm{F}}_i)) \,.
\end{split}
\end{equation}
Then, we incorporate the interacted features into the original features:
\begin{equation}
\begin{split}
\tilde{\bm{F}}_i&=\mathrm{up}(\bm{F}_i^{ca})+\bm{F}_i \,, \\
\tilde{\bm{T}}_i&=\mathrm{up}(\bm{T}_i^{ca})+\bm{T}_i \,,
\end{split}
\end{equation}
where $\mathrm{up}(\cdot)$ denotes dimension recovery implemented by deconvolution and linear layers.

\subsubsection{Feature Modulation and Multi-Scale Fusion}
Since radiology images contain multi-scale anatomical structures (e.g., lung and heart) that require model attention, we fuse the multi-scale visual features to obtain comprehensive representations. Before fusion, we modulate the visual features of different scales by interacting a global text feature $\bm{\tau}$, obtained via a projection layer in the text Transformer, with each $\tilde{\bm{F}}_{i}$ to highlight relevant regions:
\begin{equation}\label{Eq:555}
\begin{split}
\bm{M}_i&=\mathrm{MHCA}(Q=s(\tilde{\bm{F}}_i),K=\bm{\tau},V=\bm{\tau}) \,, \\
\bm{Z}&=\mathrm{Conv}_{1\times1}\circ\mathrm{Concat}(\bm{M}_1,\bm{M}_2,\bm{M}_3) \,,
\end{split}
\end{equation}
where $s$ denotes a convolutional layer to project the multi-scale features to a unified scale. $\bm{M}_{i}$ represents the modulated visual features. $\circ$ is a composition function, and $\bm{Z}\in\mathbb{R}^{C\times H\times W}$ is the fused visual feature.
\par
In addition, to incorporate spatial information into $\bm{Z}$, we concatenate it with spatial coordinates $\bm{P}\in\mathbb{R}^{2\times H\times W}$ across the channel dimension. The resulting feature is then passed through a $3\times 3$ convolutional layer to reduce the enlarged channel dimension. This porcess can be written as
\begin{equation}
\bm{X}=\mathrm{Conv}_{3\times3}\circ\mathrm{Concat}(\bm{Z},\bm{P}) \,.
\end{equation}
\par
Finally, we send $\bm{X}$ into a vision Transformer~\cite{ref_21} network such that $\bm{X}$ is transformed to a sequence of feature vectors $\{\bm{v}_{1},\bm{v}_{2},\ldots,\bm{v}_{N}\}$, where $\bm{v}_{i}\in\mathbb{R}^{D}$ for the following procedure.

\subsection{Report Decoder}
We adopt a standard Transformer decoder~\cite{ref_22} to generate reports. The decoder takes as input the adapted, fused multimodal representations from the CLIP-driven image and text encoders, and generates tokens autoregressively.

\subsection{Training and Inference}
\subsubsection{Training}
Let $\bm{I}$ be an input radiology image, and its ground truth report is denoted as $\bm{R}=\{\verb|[SOS]|,\bm{w}_{1},\bm{w}_{2},\ldots,\bm{w}_{L},\verb|[EOS]|\}$, where $\bm{w}_i\in\mathcal{V}$ represents the $i$-th token and $\mathcal{V}$ is the vocabulary set. \verb|[SOS]| and \verb|[EOS]| are the appended start and end tokens, while $L$ is the length of the sequence. At training time, we first feed $\bm{I}$ and $\{\verb|[SOS]|,\bm{w}_{1},\bm{w}_{2},\ldots,\bm{w}_{L}\}$ into the image and text encoders with our adapter to derive a multimodal representation. The Transformer decoder then takes the multimodal representation as input and $\{\verb|[SOS]|,\bm{w}_{1},\bm{w}_{2},\ldots,\bm{w}_{L}\}$ as query to generate a predicted token sequence $\{\bm{p}_{1},\bm{p}_{2},\ldots,\bm{p}_{L},\bm{p}_{L+1}\}$. We optimize the model by minimizing the cross entropy loss between the predicted sequence and the corresponding ground truth sequence $\{\bm{w}_{1},\bm{w}_{2},\ldots,\bm{w}_{L},\verb|[EOS]|\}$:
\begin{equation}
\mathcal{L}_{\mathrm{ce}}=-\frac{1}{L+1}\sum_{i=1}^{L+1}\bm{w}_i\log(\bm{p}_i) \,.
\end{equation}

\subsubsection{Inference}
During inference, our model generates texts in an autoregressive manner. Given a test image, the model is first provided an \verb|[SOS]| token as a prompt to predict the first token. The predicted first token is then concatenated with the \verb|[SOS]| token as a new prompt to predict the second token. This process continues iteratively, with the previously predicted token(s) and \verb|[SOS]| token as a prompt to predict each subsequent token, until an \verb|[EOS]| token is predicted indicating the end of generation. This autoregressive way allows the model to condition each token prediction on its previous predictions, yielding more coherent and fluent text.

\section{Experiments}

\begin{table}[!t]
    \centering
    \caption{Comparison results on the IU-Xray and MIMIC-CXR datasets. $\ast$ denotes results replicated from official code. $\dag$ indicates replicated results without pre-training on the datasets. \textbf{Bold} indicates the best results, and \underline{underline} indicates the second best results.} 
    \renewcommand\arraystretch{0.85}
    \setlength{\tabcolsep}{2pt}
    \fontsize{8.500}{11}\selectfont
    \label{table:comparison}
    \begin{tabular}{clccccccc}
    \midrule[0.85pt]
        Dataset & Method & BLEU-1 &  BLEU-2 &  BLEU-3 &  BLEU-4 &  ROUGE-L &  METEOR  \\ \midrule[0.5pt]
        \multirow{13}{*}{\rotatebox{90}{IU-Xray}} & R2Gen & 0.47 & 0.304 & 0.219 & 0.165 & 0.371 & 0.187  \\ 
        & SentSAT+KG & 0.441 & 0.291 & 0.203 & 0.147 & 0.367 & -  \\ 
        & CMCL & 0.473 & 0.305 & 0.217 & 0.162 & 0.378 & 0.186  \\
        & $\mathcal{M}^{2}$ \textsc{Tr}. \textsc{Prog.} & 0.486 & 0.317 & 0.232 & 0.173 & 0.39 & 0.192  \\ 
        & CMN & 0.475 & 0.309 & 0.222 & 0.17 & 0.375 & 0.191  \\ 
        & PPKED & 0.483 & 0.315 & 0.224 & 0.168 & 0.376 & -  \\
        & CMM+RL & 0.494 & 0.321 & \underline{0.235} & \underline{0.181} & 0.384 & 0.201  \\ 
        & XPRONET$^\ast$ & 0.491  & \underline{0.325}  & 0.228  & 0.169  & 0.387  & \underline{0.202} \\ 
        & DCL & - & - & - & 0.163 & 0.383 & 0.193 \\ 
        & M2KT & \underline{0.497} & 0.319 & 0.23 & 0.174 & \textbf{0.399} & - \\ 
        & VLCI$^\dag$ & 0.324  & 0.211  & 0.151  & 0.115  & 0.379  & 0.166 \\
        & RAMT & 0.482 & 0.31 & 0.221 & 0.165 & 0.377 & 0.195 \\ 
        & PromptMRG & 0.401 & - & - & 0.098 & 0.281 & 0.160 \\ 
        \cmidrule(r){2-8}
       &  Ours & \textbf{0.509}  & \textbf{0.349}  & \textbf{0.257}  & \textbf{0.195}  & \underline{0.395}  & \textbf{0.210}  \\ \midrule[0.5pt] 
        \multirow{12}{*}{\rotatebox{90}{MIMIC-CXR}}& R2Gen & 0.353 & 0.218 & 0.145 & 0.103 & 0.277 & 0.142 \\ 
        & CMCL & 0.344 & 0.217 & 0.14 & 0.097 & 0.281 & 0.133 \\
        & $\mathcal{M}^{2}$ \textsc{Tr}. \textsc{Prog.} & 0.378 & 0.232 & 0.154 & 0.107 & 0.272 & 0.145 \\ 
        & CMN & 0.353 & 0.218 & 0.148 & 0.106 & 0.278 & 0.142 \\ 
        & PPKED & 0.36 & 0.224 & 0.149 & 0.106 & 0.284 & 0.149 \\ 
        & CMM+RL & 0.381 & 0.232 & 0.155 & 0.109 & \underline{0.287} & 0.151 \\ 
        & XPRONET & 0.344 & 0.215 & 0.146 & 0.105 & 0.279 & 0.138 \\ 
        & DCL & - & - & - & 0.109 & 0.284 & 0.15 \\ 
        & M2KT & \underline{0.386} & \textbf{0.237} & \underline{0.157} & 0.111 & 0.274 & - \\ 
        & VLCI$^\dag$ & 0.357  & 0.216  & 0.144  & 0.103  & 0.256  & 0.136 \\
        & RAMT & 0.362 & 0.229 & 0.157 & \underline{0.113} & 0.284 & \underline{0.153} \\ 
        & PromptMRG & \textbf{0.398} & - & - & 0.112 & 0.268 & \textbf{0.157} \\ 
        \cmidrule(r){2-8}
        & Ours & 0.375  & \textbf{0.237}  & \textbf{0.165}  & \textbf{0.120}  & \textbf{0.289}  & 0.134  \\  \midrule[0.85pt]
    \end{tabular}
\end{table}

\subsection{Datasets and Evaluation Metrics}
We conduct experiments on two datasets: IU-Xray~\cite{ref_9} and MIMIC-CXR~\cite{ref_2}. IU-Xray comprises 7,470 chest X-ray images along with 3,955 radiology reports. We tokenize words with $>3$ occurrences and truncate/pad reports to 60 tokens. MIMIC-CXR is a large-scale chest X-ray dataset containing 473,057 radiographs with 206,563 associated reports. Tokens with frequency $>10$ are retained, and reports are truncated/padded to 78 tokens to conform with CLIP's specifications. For a fair and consistent evaluation on the two datasets, we use the same data splits as employed in prior works~\cite{ref_1,ref_3,ref_4,ref_5,ref_6,ref_7,ref_8,ref_100,ref_101}.

\par
We evaluate report generation quality using standard natural language processing metrics: BLEU 1-4, METEOR, and ROUGE-L. All metrics are computed with a standard evaluation toolkit~\cite{ref_26}.

\subsection{Implementation Details}
The MHSAs and MHCAs in UniCrossAdapter use 64-dim features and 4 attention heads. 
For IU-Xray, the vision Transformer and report decoder have 3 layers each, while for MIMIC-CXR, we use 6 layers due to its larger size. 
To mitigate IU-Xray's limited data, we use a consolidated vocabulary combining both datasets, enabling more diverse word projections. We choose Adam as the optimizer and use a batch size of 16 for training. We employ an initial learning rate of 1e-5 and weight decays of 5e-5 and 4e-5 for IU-Xray and MIMIC-CXR, respectively. We also apply dropout for regularization with rates of 0.09 and 0.1 for the IU-Xray and MIMIC-CXR datasets, respectively.

\subsection{Comparison with State-of-the-Art Methods}
We compare against existing methods including R2Gen~\cite{ref_1}, SentSAT+KG~\cite{ref_27}, CMCL~\cite{ref_28}, $\mathcal{M}^{2}$~\textsc{Tr}. \textsc{Progressive}~\cite{ref_29}, CMN~\cite{ref_3}, PPKED~\cite{ref_4}, CMM+RL~\cite{ref_5}, XPRONET~\cite{ref_6}, DCL~\cite{ref_8}, M2KT~\cite{ref_7}, VLCI~\cite{ref_30}, RAMT~\cite{ref_31} and PromptMRG~\cite{ref_32}.
As shown in Table~\ref{table:comparison}, the proposed approach outperforms the best competing method by 2.4\% in BLEU-2, 2.2\% in BLEU-3, 1.4\% in BLEU-4, 1.2\% in BLEU-1, and 0.8\% in METEOR on IU-Xray. While slightly lower in ROUGE-L compared to M2KT~\cite{ref_7}, our method remains the top performer overall. On the larger MIMIC-CXR dataset, our model also shows improvements of 0.8\% in BLEU-3 and 0.7\% in BLEU-4 compared to prior art, along with comparable BLEU-2 and ROUGE-L. As evidenced in previous work~\cite{ref_1,ref_3,ref_4,ref_5,ref_6,ref_7,ref_8,ref_27,ref_28,ref_29,ref_30,ref_31,ref_32}, gains on MIMIC-CXR are more marginal due to its scale. Overall, our approach achieves state-of-the-art or comparable performance on both IU-Xray and MIMIC-CXR datasets.

\begin{table}[!t]
    \centering
    \caption{Ablation results on the IU-Xray and MIMIC-CXR datasets. The best results are in \textbf{bold}. w/o denotes ``without''.}
    \renewcommand\arraystretch{0.85}
    \fontsize{8.5}{11}\selectfont
    \label{table:ablation}
    \begin{tabular}{lccccccc}
    \midrule[0.85pt]
         IU-Xray & BLEU-1 &  BLEU-2 &  BLEU-3 &  BLEU-4 &  ROUGE-L &  METEOR  \\ \midrule[0.5pt]
        w/o UniCrossAdapter & 0.302  & 0.201  & 0.146  & 0.109  & 0.375  & 0.154 \\
        w/o CLIP pre-training weights\quad\quad & 0.450  & 0.298  & 0.208  & 0.147  & 0.357  & 0.188  \\ 
       Full model & \textbf{0.509}  & \textbf{0.349}  & \textbf{0.257}  & \textbf{0.195}  & \textbf{0.395}  & \textbf{0.210}  \\ \midrule[0.5pt] 
       MIMIC-CXR & BLEU-1 &  BLEU-2 &  BLEU-3 &  BLEU-4 &  ROUGE-L &  METEOR  \\ \midrule[0.5pt]
        w/o UniCrossAdapter & 0.087  & 0.055  & 0.038  & 0.028  & 0.226  & 0.077 \\
        w/o CLIP pre-training weights & 0.351  & 0.196  & 0.118  & 0.077  & 0.250  & 0.118 \\ 
        Full model & \textbf{0.375}  & \textbf{0.237}  & \textbf{0.165}  & \textbf{0.120}  & \textbf{0.289}  & \textbf{0.134}\\
        \midrule[0.85pt]
    \end{tabular}
\end{table}

\begin{figure}[t]
\begin{center}
\includegraphics[width=\linewidth]{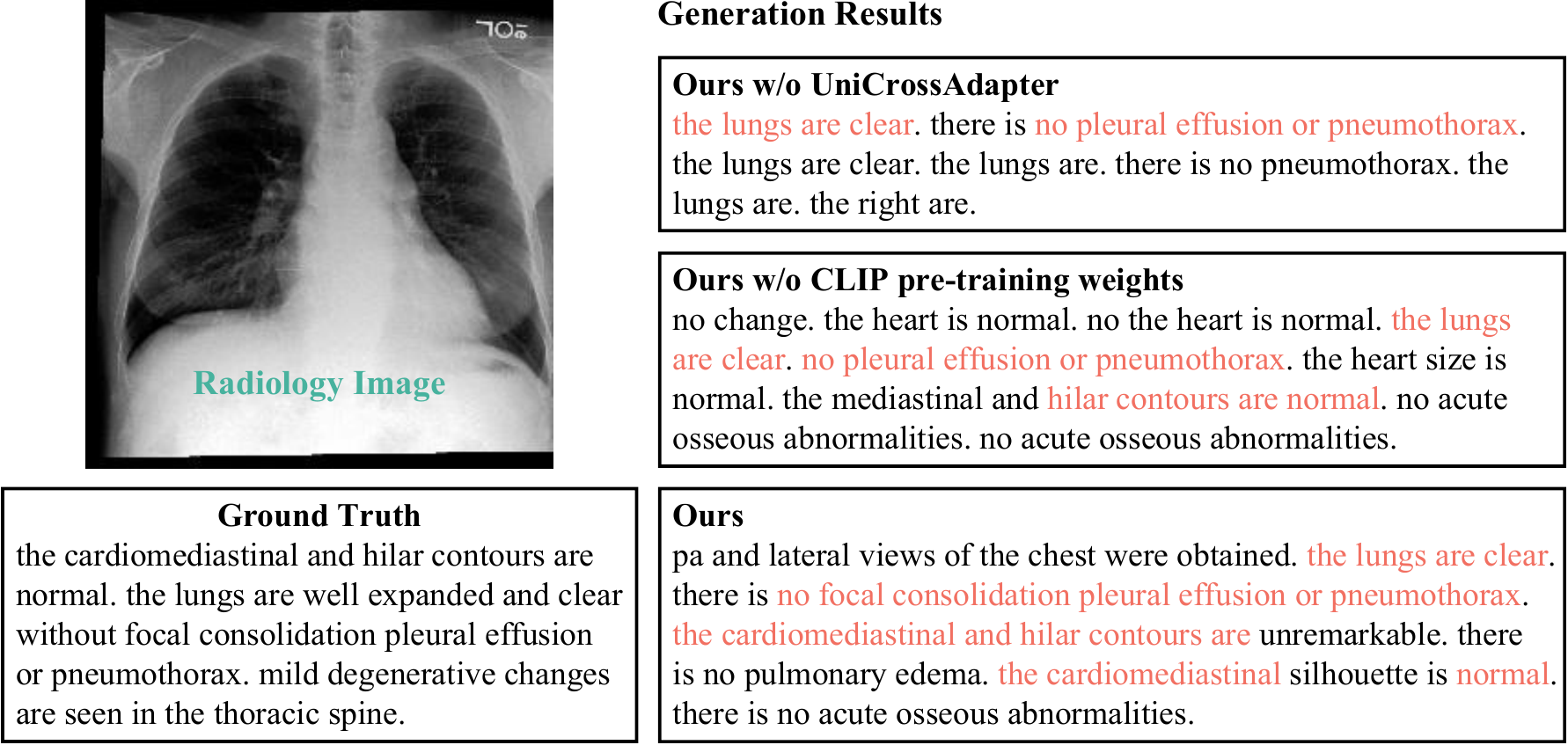}
\renewcommand{\figurename}{Fig.}
\end{center}
    \caption{Example of radiology report generation results on a test image from our model and its ablated variants. Ground truth words present in the generated reports are highlighted in color.}
\label{fig:figure_res1}
\end{figure}

\subsection{Ablation Study}
We ablate key components of our model, UniCrossAdapter and CLIP encoders, to analyze their impact quantitatively (cf. Table~\ref{table:ablation}). Removing either significantly degrades performance, validating their efficacy. This suggests that CLIP's multimodal knowledge facilitates learning cross-modal semantic alignments.

Fig.~\ref{fig:figure_res1} shows example radiology reports generated by our full model and its ablated versions. In the absence of either UniCrossAdapter or CLIP pre-training weights, the model produces noticeably inferior results. Specifically, the generated results exhibit poor grammar and a high level of repetition. This demonstrates that introducing pre-trained cross-modal knowledge from CLIP into the task of radiology report generation proves highly effective for producing more comprehensive and fluent reports. Moreover, this also highlights the significance of vision-language alignment by our adapter method for the overall model. Furthermore, we observe that reports generated by our full model demonstrate a level of professionalism comparable to ground truths.

\section{Conclusion}
In this work, we propose leveraging CLIP for the task of automated radiology report generation. Recognizing the infeasibility of fully fine-tuning such a massive model, we introduce UniCrossAdapter, a parameter-efficient fine-tuning approach to adapt CLIP to this domain. Our experiments demonstrate state-of-the-art performance on two public benchmarks. Qualitative analysis shows our model is capable of generating coherent reports describing key clinical findings in medical images. This work illustrates the promise of large pre-trained multimodal models for radiology report generation and introduces a method to make their adoption practical.

\begin{credits}
\subsubsection{\ackname} This work is supported in part by the National Key Research and Development Program of China (2022ZD0160604), in part by the Natural Science Foundation of China (62101393/62176194), in part by the High-Performance Computing Platform of YZBSTCACC, and in part by MindSpore (\url{https://www.mindspore.cn}), a new deep learning framework.

\subsubsection{\discintname}
The authors have no competing interests to declare that are relevant to the content of this paper.
\end{credits}

%
%
\bibliographystyle{splncs04}

\end{document}